\def\year{2019}\relax
\documentclass[letterpaper]{article} %DO NOT CHANGE THIS
\usepackage{aaai19}  %Required
\usepackage{times}  %Required
\usepackage{helvet}  %Required
\usepackage{courier}  %Required
\usepackage{url}  %Required
\usepackage{graphicx}  %Required
\usepackage{xcolor}
\usepackage{amsmath}
\usepackage{booktabs}
\usepackage{multirow}
\usepackage{indentfirst}
\usepackage[normalem]{ulem}
\usepackage{stfloats}

\begin{document}
\newcommand\TODO[1]{\textbf{\textcolor{red}{TODO: #1}}}
\newcommand\SH[1]{\textbf{\textcolor{purple}{#1}}}
\newcommand{\figref}[1]{Fig. \ref{#1}}
\newcommand{\adnote}[1]{\textcolor{blue}{\textbf{Anca: #1}}}
\newcommand{\adswap}[2]{\sout{#1}\textcolor{blue}{\textbf{Anca: #1}}}
\newcommand\DHM[1]{\textbf{\textcolor{green}{DHM: #1}}}

\title{Human-AI Learning Performance in Multi-Armed Bandits}
% anonymous for double-blind
%\author{Anonymous. Discipline: AI, Computer Science}
\author{Ravi Pandya, Sandy H. Huang, Dylan Hadfield-Menell, Anca D. Dragan\\
Department of Electrical Engineering and Computer Sciences\\
University of California, Berkeley\\
Berkeley, CA 94720\\
ravi.pandya@berkeley.edu, shhuang@berkeley.edu, dhm@eecs.berkeley.edu, anca@berkeley.edu
}

\maketitle

\begin{abstract}
People frequently face challenging decision-making problems in which outcomes are uncertain or unknown. Artificial intelligence (AI) algorithms exist that can outperform humans at learning such tasks. Thus, there is an opportunity for AI agents to assist people in learning these tasks more effectively. In this work, we use a multi-armed bandit as a controlled setting in which to explore this direction. We pair humans with a selection of agents and observe how well each human-agent team performs. We find that team performance can beat both human and agent performance in isolation. Interestingly, we also find that an agent's performance in isolation does not necessarily correlate with the human-agent team's performance. A drop in agent performance can lead to a disproportionately large drop in team performance, or in some settings can even \emph{improve} team performance. Pairing a human with an agent that performs slightly better than them can make them perform much better, while pairing them with an agent that performs the same can make them them perform much worse. Further, our results suggest that people have different exploration strategies and might perform better with agents that match their strategy. Overall, optimizing human-agent team performance requires going beyond optimizing agent performance, to understanding how the agent's suggestions will influence human decision-making.
\end{abstract}

\section{Introduction}
\noindent Typically, research on human-agent learning focuses on either situations where agents learn from people (e.g., learning from demonstration, recommender systems) or people learn from agents (e.g., algorithmic teaching). In contrast, our work focuses on tasks in which both the human and the agent are learning---neither is an expert yet. We are motivated by these tasks for two reasons. First, there are situations where this is obviously true, like making stock investments. But even for situations in preference learning where traditionally we assume people are experts (as in recommender systems), in reality people might actually be learning about their preferences as they take different actions and observe their outcomes. For instance, we might not know how we feel about Romanian food until we visit a restaurant, and even then we have to account for the possibility of having an unusually good or bad experience.

When it comes to learning such tasks, humans tend to struggle. We are not the best at balancing exploration and exploitation \cite{Banks1997} or internalizing our experiences thus far.
However, AI algorithms exist that can significantly outperform humans at these kinds of tasks. Thus, there is an opportunity for such agents to improve human performance by providing assistance. In particular, an agent can assist by providing \emph{suggestions}. 

\begin{figure}[t]
    \centering
    \includegraphics[width=0.9\columnwidth]{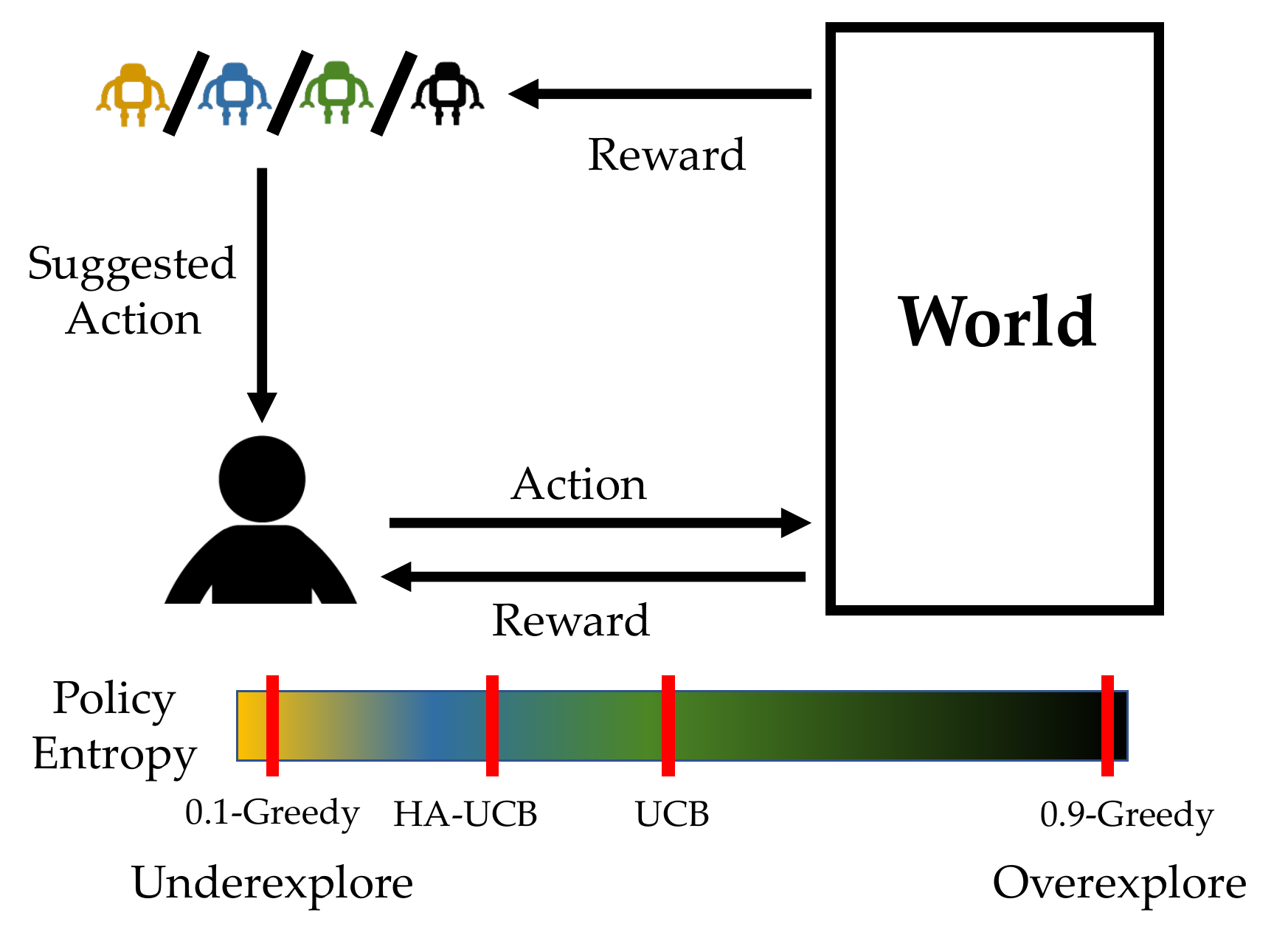}
    \caption{The suggestion-based collaborative-learning setting, in which the human takes actions in the world after seeing suggested actions from the agent.}
    \label{fig:collaboration_setup}
    \vspace{-0.2in}
\end{figure}

\noindent\textbf{Formulation of assistance.} In this work, we study the performance of human-agent teams in the context of a multi-armed bandit problem. This functions as a controlled setting in which outcomes are uncertain. Agents provide suggestions by indicating which arm the human should pull at each time step. Importantly, this setting also cleanly captures the \emph{exploration vs. exploitation} trade-off commonly encountered in reinforcement learning and robotics, in which an agent must decide whether to exploit the information it has about the world to maximize short-term reward or explore different options to ultimately find the best option. This gives us a relevant spectrum for analyzing the types of strategies employed by both humans and agents. In our experiments, we pair people with four agents spanning this spectrum (\figref{fig:collaboration_setup}).

\noindent\textbf{The team can be better than the best team member.} An encouraging result is that with the right algorithm, people not only perform better than in isolation, but they end up performing even better than the agent does in isolation. The team is more than the sum of its parts.

\noindent\textbf{Optimizing team performance is not the same as optimizing learning performance. } We expected that agent performance in isolation would correlate with human-agent team performance: the better the agent is, the better its suggestions, and thus the better the human's decisions. The first interesting result of our work is that we find evidence against this correlation. We find that a \emph{small drop} in the agent's performance can lead to a \emph{disproportionately large} drop in the human-team performance. Pairing a person with an agent that performs at their level can decrease their performance, while pairing them with an agent that is slightly better than them can increase beyond the agent's performance in isolation. Even further, a \emph{large drop} in agent performance can lead to a slight \emph{improvement} in team performance!

\noindent\textbf{Agents have implicit (rather than explicit) influence.} When analyzing how these differences in team performance came about, we were surprised to find that people were not actually changing their mind and taking the agents' suggestions. How, then, do agents influence the outcome? We found that agents actually have a more \emph{implicit} influence: suggestions do not change the person's mind immediately, but rather influence the choices the person makes \emph{later}, i.e. they change the person's strategy. Different algorithms achieve different amounts of such influence.

\noindent\textbf{People perform better with agents that are more like them.} When analyzing what might cause this difference in implicit influence, we found that people's unassisted strategies naturally group into two categories from the perspective of exploration (i.e., entropy of arm pulls over time). Each of these strategies is most similar to a specific learning algorithm, and it is that algorithm that tends to assist them best. 

Overall, our findings suggest that when using suggestions to assist people who are learning a task, we should not compute the utility of a suggestion by assuming that it will be taken. An AI's suggestions end up having different amounts of influence, largely of the more implicit kind---not changing decisions immediately, but influencing strategy over time. %We therefore need to trade off how good suggestions are with how likely they are to influence what the person does in the longer term. And because different algorithms influence people differently, it will be best to adapt the assistance to the individual person's strategy.

\section{Related Work}
\noindent 
One of the most common applications of multi-armed bandits is in recomender systems~\cite{Li2012,Li2011}. The arms of the bandit are the different options (e.g., news articles or movies) to show to the user and the reward is based on whether or not the user accepts the recommendation. This assumes the user knows what they want and the system seeks to show them options they will like. In this work, we modify this assumption: we consider a bandit setting where humans are \emph{learning} about the quality of the options themselves. Our experiments show that helping a person learn more effectively is a different task than optimizing reward in isolation. Algorithms which perform comparably on this measure can fare quite differently when used to suggest arms to a person.

Studies to understand human policies for multi-armed bandits have found that humans are suboptimal in their exploration: they over-explore \cite{Anderson2001}, under-explore \cite{Meyer1995,Horowitz_1973}, or both \cite{Banks1997}. People fall into distinct categories in terms of their cumulative regret when doing spatial bandit problems \cite{Reverdy2014}, and human performance can be well-captured using stochastic Bayesian inference algorithms \cite{Reverdy}.

Human-agent teams have been studied in resource allocation problems with uncertain outcomes, which can be thought of as multi-armed bandit problems. Prior work improved human-agent collaboration in this setting by using physical or user-interface elements~\cite{Ramchurn2015} or by modeling people's responses to an agent's \emph{actions}~\cite{Wu2015}. These scenarios differ from our work in that agents typically have full or partial control over actions, whereas we explore the setting in which only the human can take actions; the agent can only provide suggestions.

Research on improving human-robot team performance often focuses on settings in which both the human(s) and robot(s) take actions in the environment. Although our work does not deal with physical robots, successful strategies for human-robot collaboration can guide the development of successful strategies for human-AI collaboration, and vice versa. Prior work found that performance in human-robot teams improves when human and robot teammates better understand each others' actions, which is true for human-human teams as well~\cite{Marks_2002}. A robot can make itself more understandable to humans through legible motion~\cite{Dragan_2015,Stulp_2015}, increased transparency~\cite{Mercado_2016}, verbal feedback~\cite{StClair_2015}, nonverbal cues~\cite{Breazeal_2005}, revealing its incapabilities~\cite{Nikolaidis_2017,Kwon_2018}, or cross-training~\cite{Nikolaidis_2013}. These previous works on human-robot team performance typically assume a setting in which either the human or robot (or both) has ground-truth knowledge of how to perform the task optimally. In contrast, in our setting the human and agent are learning together about the task.

\begin{figure}
    \centering
    \includegraphics[width=0.8\columnwidth]{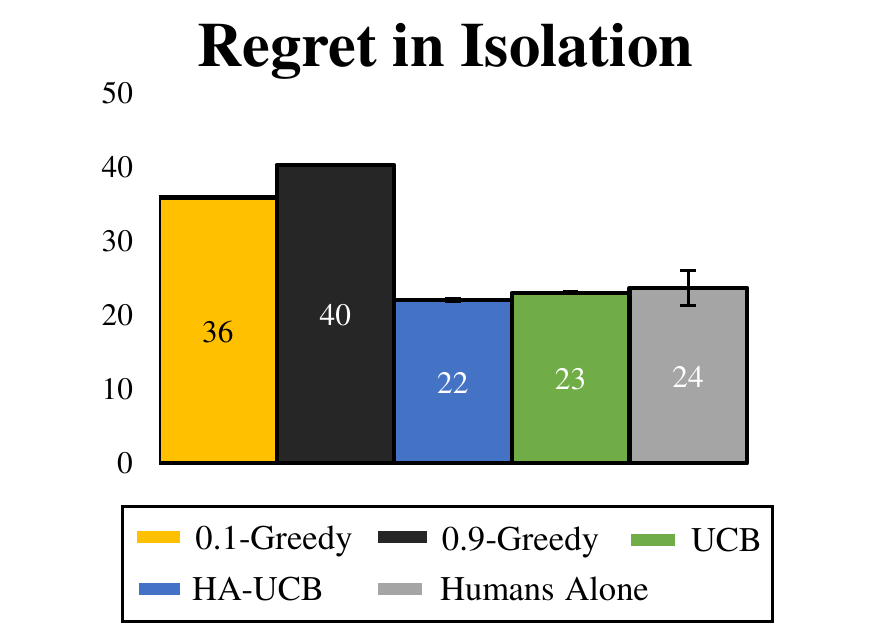}
    \caption{The average regret obtained by each agent in isolation, averaged over 10,000 simulations, and the average regret obtained by humans in isolation.}
    \label{fig:simulations}
    \vspace{-0.25in}
\end{figure}

\begin{figure}[t]
    \centering
    \includegraphics[width=\columnwidth]{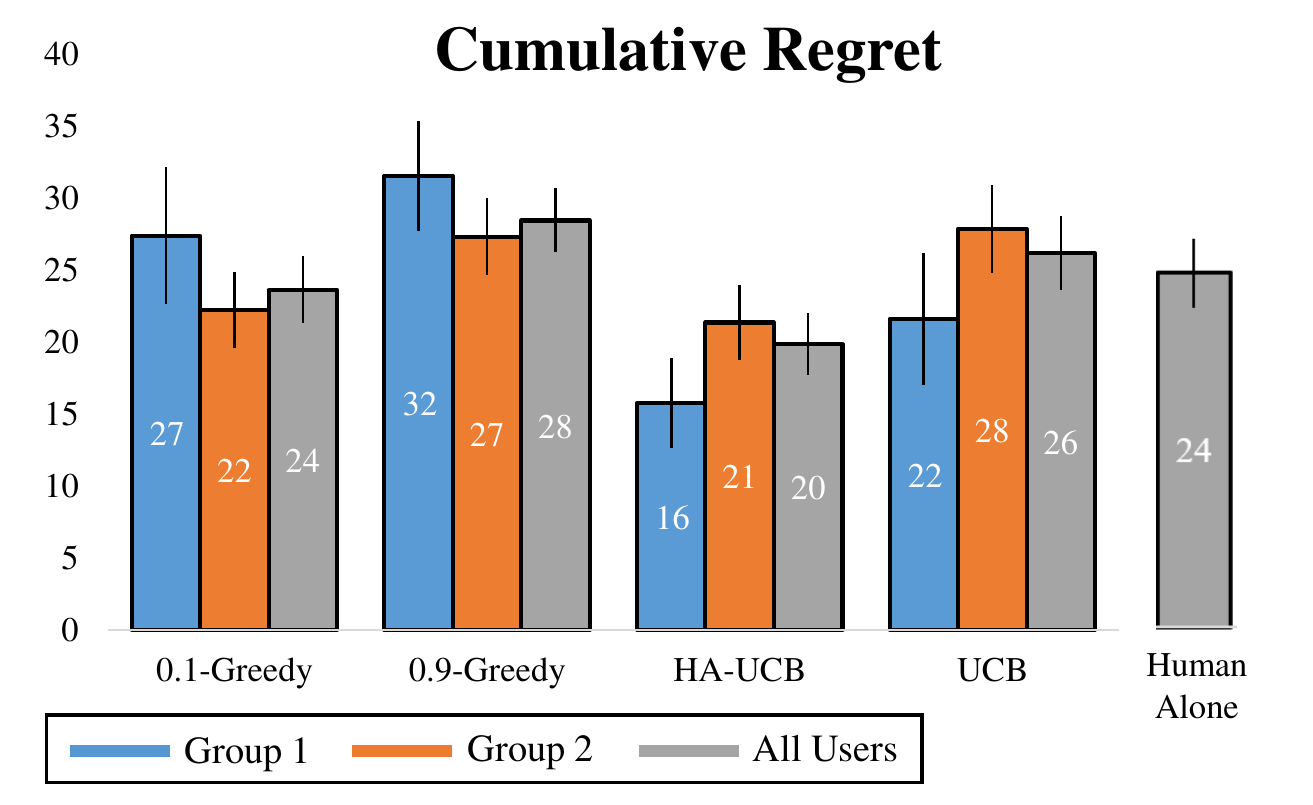}
    \caption{The average total regret accumulated when collaborating with different agents. Overall, users perform significantly better when paired with HA-UCB or 0.1-Greedy, compared to 0.9-Greedy.}
    \label{fig:total_rewards}
    \vspace{-0.2in}
\end{figure}

\section{Multi-Armed Bandit}

\begin{figure}
    \centering
    \includegraphics[width=0.98\columnwidth]{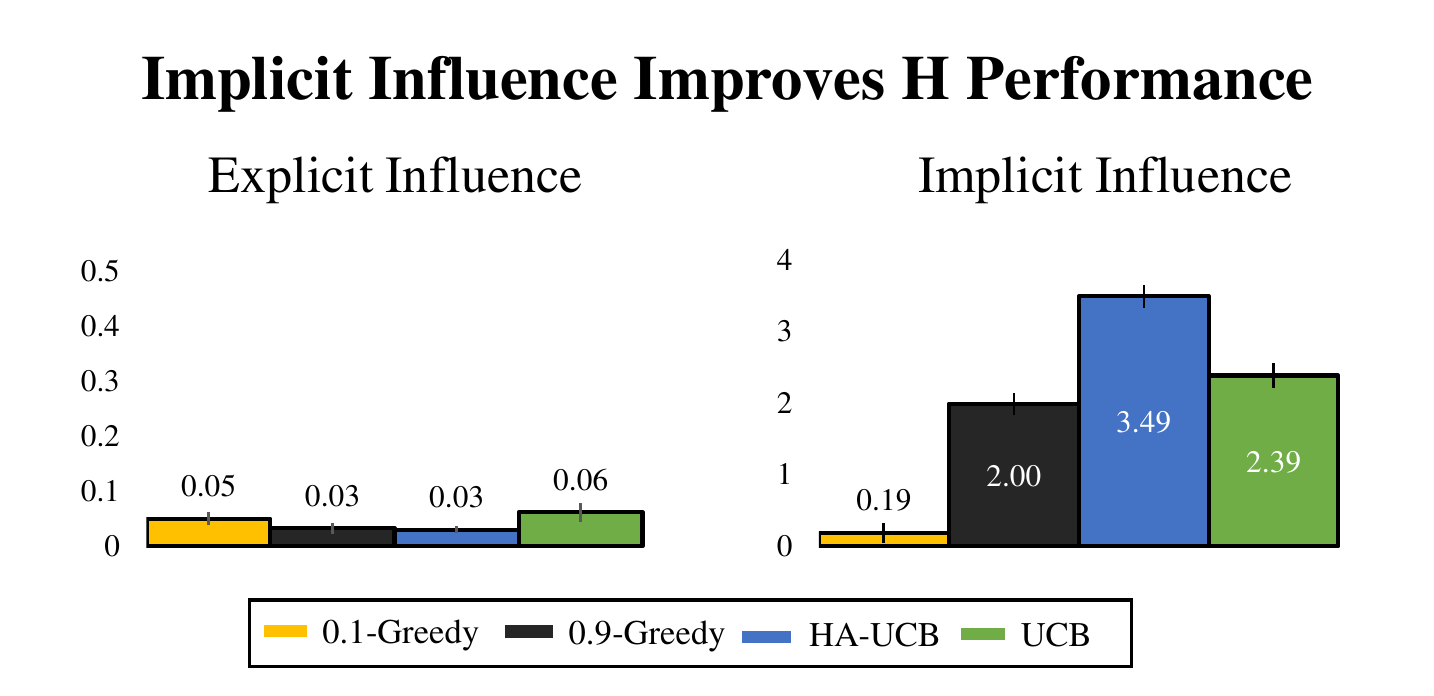}
    \caption{\textbf{Left:} \emph{Explicit influence} is the fraction of time humans immediately change their decision to the agent's suggestion when the agent suggests a different action than the human's proposed action. On average, this happens only once during the horizon of 30 pulls, so it cannot explain the increase in performance  for HA-UCB (\figref{fig:total_rewards}). \textbf{Right:} \emph{Implicit Influence} is the difference between compatibility and suggestion delay. Compatibility is how long it takes for humans in isolation to choose arms which agents would have suggested. Suggestion delay measures how long it takes for people to actually pick arms suggested by agents while being assisted by them. Here, we see HA-UCB significantly changes human strategies more than other agents, since it was initially incompatible with people, but it didn't take many iterations for people to take its suggestions.}
    \label{fig:influence}
    \vspace{-0.2in}
\end{figure}

\noindent Our user study focuses on a finite-horizon multi-armed bandit problem. A $K$-armed bandit problem is defined by random variables $X_{i,n}$ for $1 \leq i \leq K$ and $n \geq 1$. Each pull of arm $i$ yields a reward $X_{i,1},X_{i,2},...$ which are independently and identically distributed with some unknown expected value $\mu_i$. In our experiment, we used:
\begin{equation}
    X_{i,n} = \begin{cases} 
      0 & \text{w.p. } \lambda_{i,0}\\
      1 & \text{w.p. } \lambda_{i,1} \\
      2 & \text{w.p. } \lambda_{i,2} \\
      3 & \text{w.p. } \lambda_{i,3} \\
      4 & \text{w.p. } \lambda_{i,4} \\
   \end{cases}
\end{equation}
where $\lambda_i$ is different and fixed over time for each arm, and $\sum_{j=0}^4 \lambda_{i,j} = 1$.

Intuitively, since the arms are unchanged over time, the best strategy for maximizing reward is to always choose the arm $i$ which has the highest expected reward, $\mu_i = \lambda_i$ [0 1 2 3 4]$^T$, where $\lambda_i$ is the row vector of all probabilities for arm $i$. Thus, we can evaluate different \emph{policies}, or \emph{allocation strategies}, based on how much worse they are doing than this optimal policy. If we define $T_i(n)$ as total reward gained from arm $i$ in the first $n$ plays, then we can define the \emph{regret} for some policy after $n$ pulls as
\begin{equation}
    \mu^{*}n - \sum_{j=1}^K T_j(n) \text{ where } \mu^{*}=\max_{1 \leq i \leq K} \mu_i
\end{equation}
The goal is to find a policy which minimizes the total regret.

Since we have no prior knowledge of which arm may be the best, this introduces a classic \emph{exploration vs. exploitation} trade-off as previously noted. Many policies for this problem have been explored; among these, the Upper Confidence Bound (UCB) algorithm is simple and has a bound on expected regret that is logarithmic in the number of pulls \cite{Auer2002}.

\subsection{UCB Policy}
This policy starts by sampling each arm once. On each subsequent pull $n$, it picks the arm $i$ that maximizes 
\begin{equation}
    \bar \mu_i + \sqrt{\frac{2\ln n}{n_i}},
\end{equation}
where $\bar \mu_i$ is the average reward obtained from arm $i$ and $n_i$ is the number of times that arm has been played so far.

\subsection{Horizon-Aware UCB Policy}
While UCB has a good \emph{asymptotic} bound on its regret, it is possible for other agents to outperform it in short time horizon problems, like the ones in our user studies. 

One way to improve the performance of UCB on such problems is to make it more greedy. Note that the vanilla UCB algorithm always takes the arm which maximizes the sum of two terms. The first term, $\bar \mu_i$, represents exploitation or greediness, since it is the average reward seen from arm $i$ so far. The second term, $\sqrt{(2\ln n) / n_i}$, represents how confident the algorithm is in its estimate of $\mu_i$, which can be thought of as an exploration term. Thus, to make UCB more greedy, we can reduce the magnitude of this exploration term.

In particular, we introduce a parameter $\gamma$ which starts at 1 on the first pull and linearly decays to 0 for the last pull of the time horizon. On each pull $n$, this new Horizon-Aware UCB (HA-UCB) policy now picks the arm $i$ that maximizes 
\begin{equation}
    \bar \mu_i + \gamma \sqrt{\frac{2\ln n}{n_i}}.
\end{equation}

Decaying $\gamma$ in this way means a HA-UCB policy will act exactly like UCB on the first pull and exactly like a perfectly greedy agent on the last pull. HA-UCB slightly outperforms UCB on the tasks we consider (\figref{fig:simulations}).

\subsection{$\epsilon$-Greedy Policy}
$\epsilon$-greedy policies pick the arm with the highest average reward so far with probability 1-$\epsilon$ and pick a random arm with probability $\epsilon$. We include one over-exploring agent (0.9-Greedy) and one under-exploring agent (0.1-Greedy). In isolation, 0.1-Greedy obtains lower regret than 0.9-Greedy, but both have significantly higher regret than UCB and HA-UCB (\figref{fig:simulations}).

\section{Experimental Design}
\subsection{User Study}
We ran a user study in which participants played a game with multiple slot machines (i.e., arms). Users collaborated with different agents, that suggest which slot to pick at each iteration. We introduced these agents as ``robots'' to users, to concisely communicate that the suggestions were from a non-human actor. 

When collaborating with an agent, users are first asked which slot they would like to play before seeing the agent's suggestion. After this, the user is shown the agent's suggestion via highlighting the slot(s) the agent would pick. If the agent has no preference among multiple slots (for instance, when greedy algorithms choose randomly), then all of those slots will be highlighted. Once they see the suggestion, users are free to select any slot. 

\subsection{Manipulated Variables}
We manipulated the \emph{learning algorithm} with five levels: \emph{Unassisted}, \emph{0.1-Greedy}, \emph{0.9-Greedy}, \emph{UCB}, and \emph{HA-UCB}. We purposefully chose these agents to span the \emph{exploration vs. exploitation} spectrum. We used a within-subjects design for this variable and counterbalanced the order. 

\subsection{Objective Measures}
\begin{itemize}
    \item \textbf{Regret:} The total regret accumulated after all $n=30$ pulls.
    \item \textbf{Inherent Compatibility:} The amount of time it takes for users in isolation to pick arms which each agent \emph{would have} suggested had they been assisting.
    \item \textbf{Explicit Influence:} The percentage of time the human's choice changes to the agent's suggestion after seeing it.
    \item \textbf{Implicit Influence:} The difference between inherent compatibility and how long users \emph{actually} take to pick arms that agents suggest.
    \item \textbf{Decision Scores:} The normalized score assigned by HA-UCB to the decision made, based on the history of pulls and rewards. This allows us to analyze users' strategies before and after getting assistance.
    \item \textbf{Entropy:} The entropy of the distribution over how often users choose each arm; this measures whether they under-explore or over-explore. A perfectly greedy policy will have entropy 0, whereas a perfectly uniform policy (with $K=6$ arms as in the user study) will have entropy 2.58.
\end{itemize}

\subsection{Subjective Measures}
We also care about the users' perceptions of the agents, so we ask three Likert scale questions about whether they \emph{trusted} the agent, whether they thought the agent was \emph{useful}, and whether they followed the agent's \emph{advice}. We also ask users to \emph{rank} the agents in order of how much they enjoyed collaborating with them.

\subsection{Participants}
We used Amazon Mechanical Turk to recruit a total of 52 users (33\% female, mean age 33).
Users were compensated $\$$3.75 for the study, which lasted approximately 20 minutes. Users were also given up to a $\$$1 reward depending on their average payout across all collaboration settings. Users were informed of this reward bonus before starting the study, in order to incentivize them to pay attention and try their best. 

\section{Analysis}

{\begin{table*}[t]
    \centering
    \begin{tabular}{lcccc}
        \toprule \hline
        \textbf{Statement} & \textbf{0.1-Greedy} & \textbf{0.9-Greedy} &\textbf{HA-UCB} & \textbf{UCB}\\
        ``I trusted the agent'' & 4.2 $\pm$ 0.23 & 2.5 $\pm$ 0.20 & \textbf{4.8 $\pm$ 0.21} & 4.2 $\pm$ 0.25\\
        \midrule
        ``I thought the agent was useful'' & 4.1 $\pm$ 0.25 & 2.6 $\pm$ 0.22 & \textbf{4.8 $\pm$ 0.24} & 4.2 $\pm$ 0.27\\
        \midrule
        ``I followed the agent's advice'' & 3.8 $\pm$ 0.25 & 3.5 $\pm$ 0.23 & \textbf{4.7 $\pm$ 0.25} & 4.0 $\pm$ 0.27\\
        \midrule
        Rank [1: best, 4: worst] & 2.5 $\pm$ 0.14 & 3.0 $\pm$ 0.13 & \bf{2.0 $\pm$ 0.14} & 2.5 $\pm$ 0.15\\
        \bottomrule
    \end{tabular}
    \caption{Post-study Likert ratings. Users prefer to work with HA-UCB significantly more than with other agents.}
    \label{tab:subjective_measures}
    \vspace{-0.1in}
\end{table*}
}

\noindent\textbf{The team can be better than the best team member.} In isolation, UCB and HA-UCB perform the best in terms of cumulative regret, scoring 23 and 22 respectively. Humans in isolation perform similarly, getting an average regret of 23. $\epsilon$-Greedy agents perform notably worse than this, with 0.1-Greedy and 0.9-Greedy getting regret of 36 and 40 respectively (\figref{fig:simulations}). 

The performance of the human-agent team only improves when people are paired with HA-UCB. This is expected, but it is exciting to see that the team \emph{outperforms} HA-UCB in isolation (\figref{fig:total_rewards}). Not only are people able to improve their own performance, but the human-agent team---when paired with the right agent---can do better than either humans or agents in isolation. Interestingly, we find that particular individuals perform even better still (significantly) when paired with HA-UCB; these are the individuals labeled ``Group 1'', which we will discuss later in this section. 

\noindent\textbf{Optimizing team performance is not the same as optimizing learning performance.} We expected that in general, the performance of the agent in isolation will have some correlation with the performance of the human-agent team. But even though the best agent led to the best team, the correlation did not hold in general.

We ran repeated measures ANOVA for our objective measures of Regret and did post-hoc analyses with Tukey HSD. We found that the learning algorithm factor has a significant effect on Regret ($F(3,48)=11.529,p<0.01$) and that HA-UCB is significantly different from 0.9-Greedy (Tukey HSD).
One interesting result is that while HA-UCB only outperforms UCB by 1 point in isolation, human-HA-UCB teams significantly outperform human-UCB teams (\figref{fig:total_rewards}), by an average of 6 points. 
This sixfold increase in the difference in performance indicates that HA-UCB's suggestions were somehow more helpful and made more sense to people than UCB's, leading to people being able to make much more informed decisions. 

Another interesting result is that the team's performance can improve slightly (or at least remain unaffected) despite pairing the human with a \emph{worse} agent. UCB in isolation outperforms .1-Greedy (\figref{fig:simulations}) by an average of 13 points, while the two human-agent teams perform very similarly. 0.1-Greedy even gets slightly lower regret by 2 points on average (\figref{fig:total_rewards}). Though this difference is not statistically significant, it stands in stark contrast to how much better UCB performs than 0.1-Greedy in isolation.

Despite team performance not correlating to agent performance in isolation, the users' ratings did. Table \ref{tab:subjective_measures} shows the subjective measures: agents better in isolation are rated higher.

These results indicate that simply improving an agent's isolated performance does not correlate with improving its ability to assist humans. Assistance is more subtle, and we explore what influences human-team performance in the remainder of this section.

\begin{figure}[t!]
    \centering
    \includegraphics[width=\columnwidth]{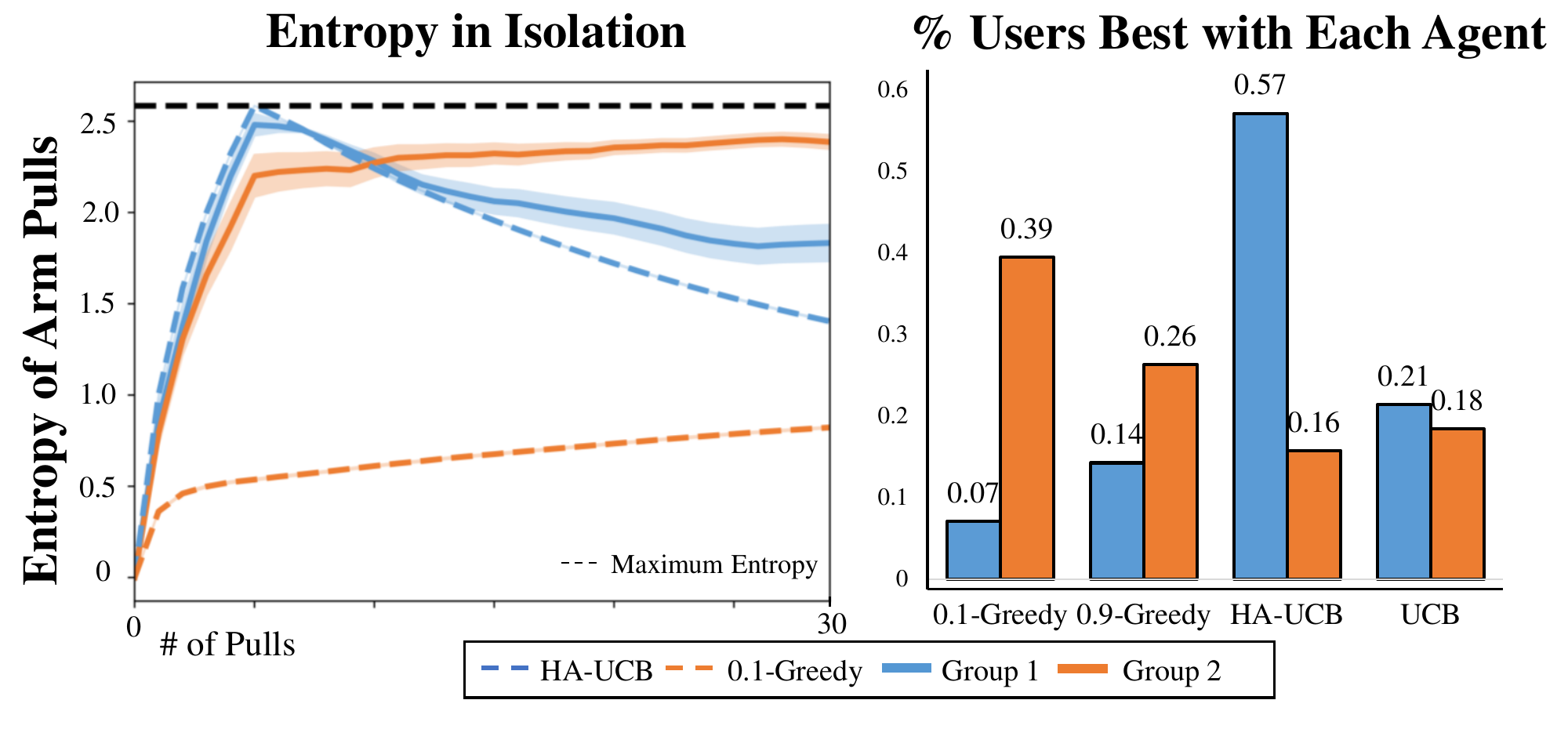}
    \caption{\textbf{Left:} The entropy of the distribution over how often users and agents pick each arm in isolation as a function of the number of pulls. Group 1 explores more fully initially and then becomes greedy, Group 2 explores at the same rate throughout. Curves for agents are averaged over 10,000 trials. HA-UCB has a similar shape to Group 1 and 0.1-Greedy has a similar shape to Group 2. \textbf{Right:} The percentage of users who obtain their highest score with each agent. Most users in Group 1 get their highest score with HA-UCB, while nearly 40$\%$ of users in Group 2 get their highest score with 0.1-Greedy. The agents' exploration strategies align with the users' strategies in these cases, which seems to improve team performance.}
    \label{fig:entropy_curve_groups}
    \vspace{-0.2in}
\end{figure}

\noindent\textbf{Agents have implicit (rather than explicit) influence.} One hypothesis for what causes the difference in team performance is that agents convince people to change their mind, going against their initial choice for an arm. However, when we measure how often users change their mind to agree with the agent's suggestion after seeing it, we found no influence from the agents. Overall, the percent of time people are explicitly influenced like this is close to 4$\%$ across all agents, which corresponds to approximately one decision over the user's 30-decision time horizon (\figref{fig:influence}). It seems unlikely that this one decision could affect the performance of human-agent teams so drastically. But if people are not changing their decisions to match the agent's suggestions, how are different agents leading to different team performances? 

What we found is that users' initial choices are different when interacting with the different agents: there is an \emph{implicit} influence that agents have, whereby the suggestions agents make at one time point do not lead the person to explicitly change their mind about which arm they pull now, but affect their choices in the future.

We measure this by looking instead at how long it takes until people actually follow agents' suggestions (\figref{fig:influence}). For each agent, we look at how many pulls it takes each user on average to ultimately pick the arms it suggests. We call this the \emph{delay} in accepting the agent's suggestions. This is an informative measure, but to understand the agent's influence, we need to compare this to some baseline to understand how much users' decisions are actually changing. To do this, we simulate the agents making suggestions based on people's decisions in isolation and similarly measure how long it takes for people to choose arms which the agents would have suggested. We refer to this as a user's \emph{compatibility} with each agent, since it tells us how quickly users would have taken agents' suggestions regardless. We refer to the difference between the delay in accepting the agent's suggestions and the human's compatibility with that agent as the agent's \emph{implicit influence}. 

We ran repeated measures ANOVA for our objective measures of Regret and did post-hoc analyses with Tukey HSD. We found significant effects for the learning algorithm on Implicit Influence ($F(3,48)=122.04,p<0.0001$). HA-UCB had higher influence than UCB and 0.1-Greedy (Tukey HSD). Finally, we found significant effects for the agent assistance factor (assisted or unassisted) on Implicit Influence ($F(1,48)=13.45,p=0.0004$). 

When not assisted by any agent, people are most similar to the 0.1-Greedy agent, only having about a 2 pull delay before they choose what it would have suggested. In contrast, users take around 5 or 6 pulls before picking what the other agents would have suggested. When people are actually seeing agents' suggestions, this average falls across the board. This tells us that although agents do not explicitly change users' minds, their suggestions have an implicit influence on their strategies going forward.

When viewed from this lens, we do actually see a difference in influence between the agents! The time it takes for users to take HA-UCB's suggested arms goes down by the largest fraction of all the agents. With no assistance, people take on average 5.4 pulls before taking what HA-UCB would have suggested, whereas people actually take less than 2 pulls on average before pulling arms which it actually suggested. In contrast, there is little difference here for 0.1-Greedy between the unassisted and assisted settings: people take about 2.5 pulls before following its decisions in both cases (\figref{fig:influence}). The other agents (UCB and 0.9-Greedy) see a slight improvement, but not nearly as large as that of HA-UCB. 

Though we see that HA-UCB has the most implicit influence on people, they do not directly take its suggestions very often. Instead, they are influenced to change their overall strategy after seeing its suggestions.

\noindent\textbf{People  perform  better  with  agents  that  are  more  like them.}  Next, we turned to understanding what might be responsible for this difference in influence. We looked at unassisted users first and plotted the entropy of the distribution of arms they had selected up to each time step. We found that users naturally fell into two distinct groups. Users were manually separated into these groups and no users were excluded. 21 users were placed into Group 1 and 31 into Group 2.

As we see in \figref{fig:entropy_curve_groups}, people in Group 1 will initially explore all or almost all arms as evident by the entropy steadily increasing to the maximum entropy 2.58. The entropy then steadily decreases, indicating that users will settle on one or two arms to continue pulling for the remainder of the time horizon. In contrast, Group 2 will continue to explore all arms at approximately the same rate for the entirety of the time horizon, as evident by the entropy curve increasing to and leveling out at 2.4. 

Remarkably, if we look at which agent each person in these groups got their personal high score with, there is a distinct difference between them. As shown in \figref{fig:entropy_curve_groups}, a majority of users in Group 1 (57$\%$) get their highest score when collaborating with HA-UCB, whereas only only 16$\%$ of users in Group 2 do. In contrast, only 7$\%$ of users in Group 1 get their highest score when collaborating with 0.1-Greedy as compared with 39$\%$ of users in Group 2. Now, if we plot the entropy curves for these two agents in isolation averaged over 10,000 trials (\figref{fig:entropy_curve_groups}), we see the shapes tend to correspond with those of the two groups. This lends credence to the idea that users perform best when being assisted by an agent which acts like them. 

We see that Group 1, which matches HA-UCB, reduces regret to 16 when assisted by HA-UCB, which is far lower than even HA-UCB's performance (regret 22). While we do not make statistical claims, people have a better sense of what the best arm looks like (in terms of average reward), whereas HA-UCB starts with no information. With HA-UCB assisting, people are inclined to explore arms which they would not have on their own, so the team in total is better able to identify the best arm than either would have in isolation. The same group performs worse with 0.1-Greedy, decreasing performance also when compared to how well these users did in isolation. Particularly surprising is that more Group 2 users perform better with 0.1-Greedy than with HA-UCB, despite HA-UCB being the better algorithm. Group 2 performs slightly better when assisted by 0.1-Greedy than when unassisted, whereas Group 1 performs worse.\footnote{On the surface, this analysis seems to contradict that the average total regret accumulated by Group 2 when collaborating with HA-UCB is not significantly different from when collaborating with 0.1-Greedy (\figref{fig:total_rewards}). However, we notice that when users in Group 2 get lower regret with 0.1-Greedy than with HA-UCB it is by an average of 13 points, whereas those that get lower regret with HA-UCB do better by an average of 19 points. Thus it is the case that while more users perform better with 0.1-Greedy, the average regret from collaborating with the two agents is approximately the same. }

{\begin{table}[t]
    \centering
    \begin{tabular}{lcc}
        \toprule \hline
         \textbf{Statement} & \textbf{Group 1} & \textbf{Group 2} \\
         ``I trusted the agent'' & 3.4 $\pm$ 0.35 & \textbf{4.5 $\pm$ 0.27}\\
        \midrule
        ``I thought the agent was useful'' & 3.2 $\pm$ 0.47 & \textbf{4.4 $\pm$ 0.28}\\
        \midrule
        ``I followed the agent's advice'' & 2.7 $\pm$ 0.42 & \textbf{4.2 $\pm$ 0.28}\\
        \midrule
        Rank [1: best, 4: worst] & 3.0 $\pm$ 0.25 & \bf{2.3 $\pm$ 0.17}\\
        \bottomrule
    \end{tabular}
    \caption{Post-study Likert ratings for 0.1-Greedy. (Differences between the two groups were negligible for other agents.) Group 2, who performs better when collaborating with 0.1-Greedy, overall has a positive view of the agent while Group 1 has a negative view.}
    \label{tab:greedy_subjective_measures}
    \vspace{-0.2in}
\end{table}
}

When looking at the subjective measures split by groups, we find that they disagree in their opinion of 0.1-Greedy. Group 1, who is more aligned with HA-UCB, rates the 0.1-Greedy much lower than Group 2 (Table \ref{tab:greedy_subjective_measures}).

Overall, we find that people have different strategies, and many of them team up best with agents that match their strategy. We found it striking that 39\% of people with the greedy-like strategy perform best with greedy, whereas only 16\% of them perform best with HA-UCB, and this is in spite of HA-UCB's superiority in isolation. 

\section{Discussion and Future Work}
\noindent We saw that human-agent teams can outperform humans and agents in isolation. But our analysis suggests that achieving this, or even just improving upon human performance, is much more subtle than we expected. The agent's suggestions do not change a person's decisions explicitly, but rather influence their later decisions. Further, people benefit differently from different agents, depending on the similarity between their strategy and the agent's.

These results show that helping a person manage exploration exploitation trade-offs is distinct from directly making those trade-offs. We can alternatively formulate this problem as a cooperative game between the human and the robot~\cite{Hadfield2016}, where both the robot and the human are optimizing to maximize the cumulative reward from the \emph{human's} arm selections. Crucially, the robot is forced to operate through making changes to the human's internal or information state. In future work, we plan to explore this formulation of the problem and develop algorithms that leverage models of human internal state to make helpful suggestions and work with humans to explore and exploit appropriately to maximize long term reward.

\bibliographystyle{aaai} \bibliography{references}

\clearpage
\begin{figure*}[t!]
    \centering
    \includegraphics[width=\textwidth]{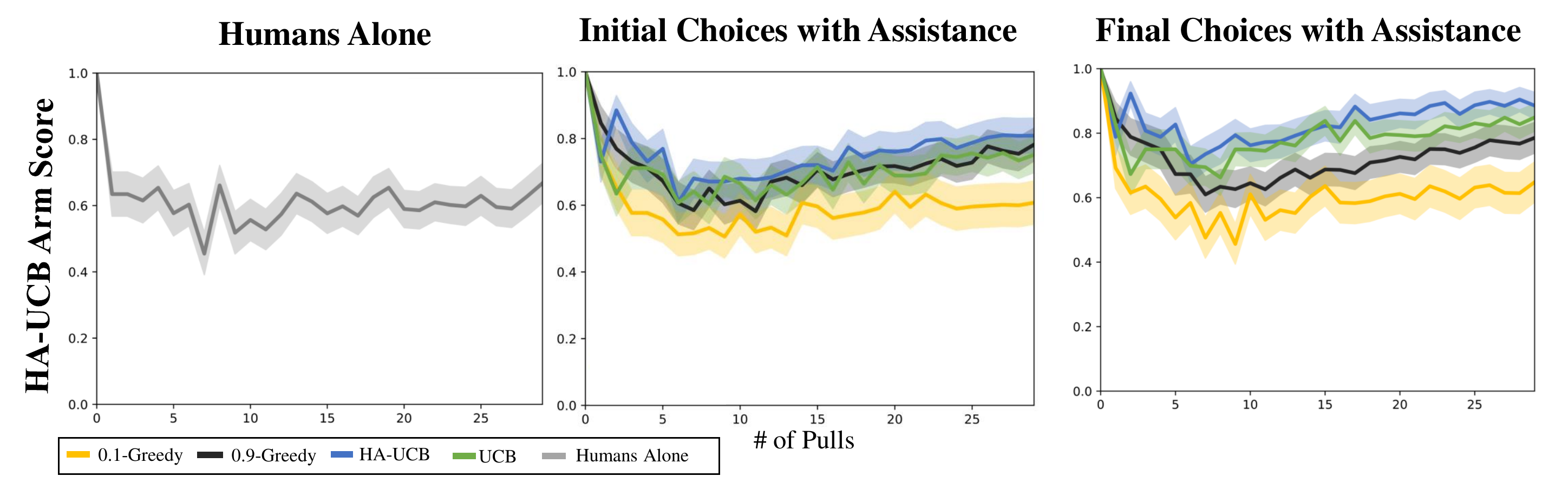}
    \caption[width=\textwidth]{The y-axis shows the normalized score assigned by HA-UCB to the decision made, based on the history of pulls and rewards. \textbf{Left:} The average scores obtained by human strategies in isolation. \textbf{Middle:} The average scores obtained from humans' initial choices while collaborating with each agent. \textbf{Right:} The averages scores obtained from humans' final choices while collaborating with each agent. Note that the scores are highest for HA-UCB and that the curves are different for people when collaborating with an agent versus in isolation, meaning that their strategies are altered.}
    \label{fig:assistance_strategies}
\end{figure*}
\section*{Appendix: Strategies With and Without Assistance}
A second way we analyzed implicit influence is by looking at the (normalized) score that HA-UCB assigns to the users' initial and final choices given the history of pulls (\figref{fig:assistance_strategies}). Though the scores people get here do not necessarily correlate with their regret, we can use this to measure whether human strategies are changing. If strategies were not being influenced by agents, we should expect these scores to stay approximately the same. However, we see that the curves for humans in isolation are significantly different from those when collaborating with agents. Again, this tells us that agents influence people's decisions implicitly rather than explicitly.

\end{document}